\documentclass[10pt,twocolumn, letterpaper]{article}

\usepackage{times}
\usepackage{epsfig}
\usepackage{graphicx}
\usepackage{amsmath}
\usepackage{amssymb}
\usepackage{array}
\usepackage{mathrsfs}
\usepackage{algorithm, algpseudocode}

\usepackage[utf8]{inputenc}
\usepackage[left=0.7in,right=0.7in,top=0.7in,bottom=1.0in]{geometry}

\usepackage[english, french]{babel}                                                         

\usepackage{bm}
\usepackage{siunitx}

\newcommand{\ssect}{\subsection}

\newcommand*\mean[1]{\bar{#1}}

\usepackage{algpseudocode,algorithm,algorithmicx}

\algrenewcommand\algorithmicrequire{\textbf{Precondition:}}
\algrenewcommand\algorithmicensure{\textbf{Postcondition:}}

\algdef{SE}[SUBALG]{Indent}{EndIndent}{}{\algorithmicend\ }%
\algtext*{Indent}
\algtext*{EndIndent}

\title{Event-Based Features Selection and Tracking from Intertwined Estimation of Velocity and Generative Contours }

\author{Laurent Dardelet\\
Sorbonne Universités\\
{\tt\small dardelet.l@gmail.com}
\and
Sio-Hoi Ieng\\
Sorbonne Universités\\
{\tt\small sio-hoi.ieng@upmc.fr}
\and 
Ryad Benosman\\
University of Pittsburgh\\
Carnegie Mellon University\\
Sorbonne Universités\\
{\tt\small benosman@pitt.edu}
}

\begin{document}
\maketitle

\makeatletter
\makeatother


\section*{Abstract}

This paper presents a new event-based method for detecting and tracking features from the output of an event-based camera. Unlike many tracking algorithms from the computer vision community, this process does not aim for particular predefined shapes such as corners. It relies on a dual intertwined iterative continuous -pure event-based- estimation of the velocity vector and a bayesian description of the generative feature contours. By projecting along estimated speeds updated for each incoming event it is possible to identify and determine the spatial location and generative contour of the tracked feature while iteratively updating the estimation of the velocity vector. Results on several environments are shown taking into account large variations in terms of luminosity, speed, nature and size of the tracked features. The usage of speed instead of positions allows for a much faster feedback allowing for very fast convergence rates. 

\section{Introduction}

Although event-based cameras are becoming popular and available, there is currently no solution to efficiently select and track features from the output of these sensors in a pure event-based manner. By pure event-based, we mean an iterative computation process where every incoming event adds a small contribution to the global computation that needs to be carried out without requiring the use of ''temporal frames'' to operate on large portions of time. 
The problem of feature detection and tracking is challenging when dealing with an event-based sensor as there is the constant temptation of looking at what the community is used to and build frames out of events to be able to reuse old concepts ~\cite{HarrisCorner,Rosten2006}. These solutions are rarely providing satisfactory performances when compared to results reported by conventional frame-based methodology that provide both absolute light measurement and higher spatial resolution. Considering event-based camera implies making full use of its properties of low redundancy and high temporal accuracy that are both lost if one considers portions of time to process incoming event-based information. This paper introduces an event-based solution for efficiently selecting and tacking features from the output of an event-based camera. Considering a pure event-based methodology allows to write the problem in the time domain as an inter-dependent process between detection, tracking and continuous velocity estimation. The method does not assume any priors as it is often the case in standard image processing.\\
 Event-based vision sensors are gaining popularity within the conventional computer vision community as they are offering many advantages that frame-based cameras are not able to provide without an unreasonable increase in computational resources. Low computation needs is especially achieved by a lower redundant data acquisition, while presenting lower latency than standard cameras, achieved via a highly precise temporal and asynchronous level crossing sampling. This level crossing sampling, opposed to the fixed frequency sampling implemented in standard cameras, is highly valuable in robotics for example, requiring agility and fast reactions in situations such as drone navigation or autonomous driving.
\begin{figure}[h]
    \centering
    \includegraphics[width=\columnwidth]{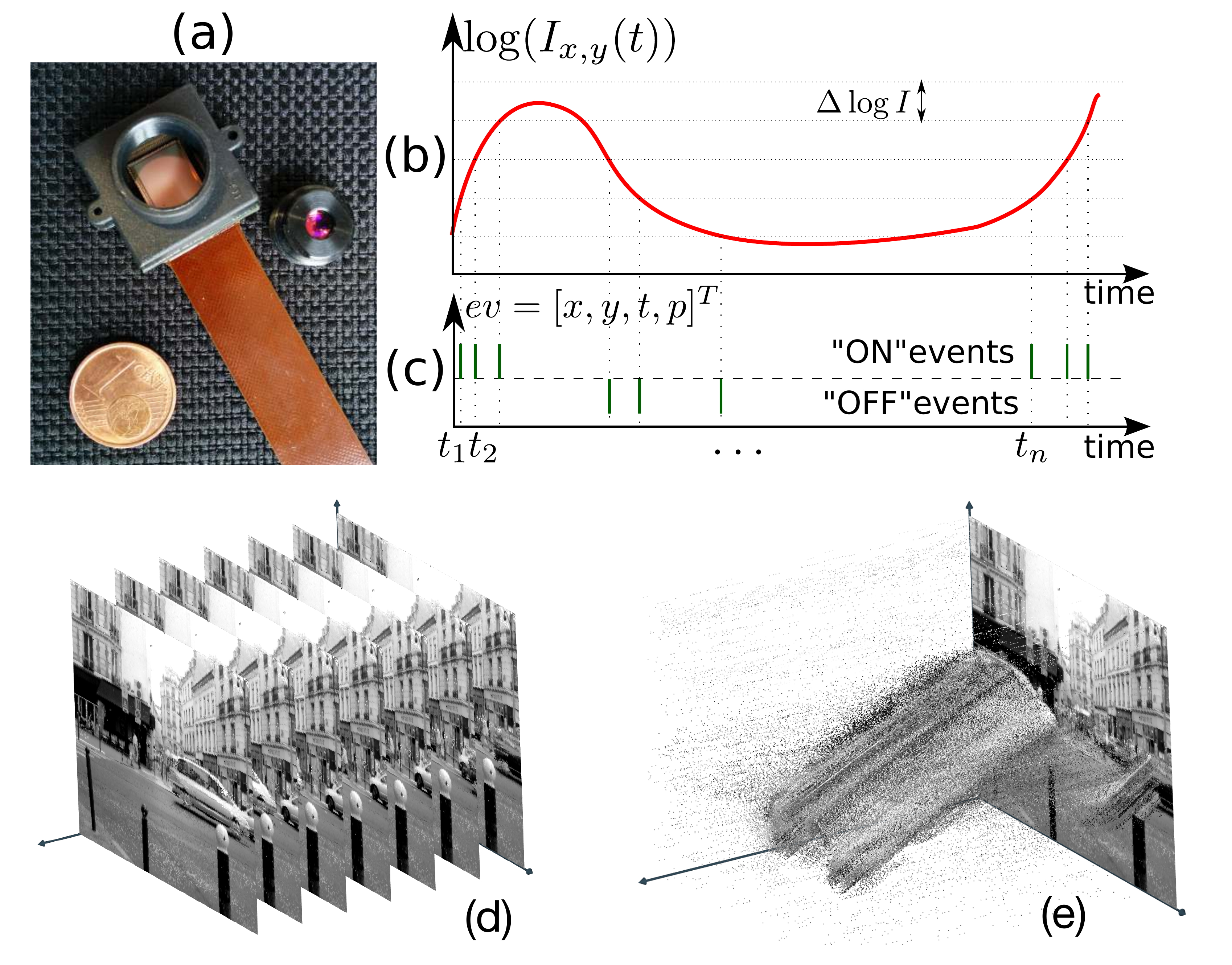}
    \caption{Illustration of event-based encoding of visual signals. (a) Event-based camera used~\cite{Posch_2011,Guo2017}. (b) Log of the luminance of a pixel located at $[x, y]^T$ (c) Asynchronous temporal contrast events generated with respect to the predefined threshold  log(I) (d) the output of a conventional frame based cameras at some 30-60Hz (d) high temporal precision event-based output at around (1$\mu$s) providing an almost continuous output compared to the scenes dynamics.\label{fig_lvlcrossing}}
\end{figure}
Most of readily available event-based vision sensors stem from the Dynamic Vision Sensor (DVS)~\cite{Lichsteiner2008}. As such, they work in a similar manner by capturing relative luminance changes: each individual pixel emits an "event" each time the luminance crosses a predefined threshold at $\mu$s precision. An event contains both the pixel coordinates from which it originates and the time it is triggered, hence it is usually given defined as $e=(\bm{X},t)$. Additional information can be added such as the sign of that change as illustrated in Figure~\ref{fig_lvlcrossing}(b) and conventionally referred to as "ON/OFF" polarities. Other variations of the silicon retinas implement functions such as capturing absolute luminance information with the same asynchronous philosophy~\cite{Posch_2011,Guo2017} not used in this work or by implementing an hybrid solution that captures grey scale frames as in a regular camera~\cite{Brandli2014}. Thanks to the asynchronous level crossing sampling, all of these sensors are able to reach over 120dB in dynamic range. This asynchronous visual information acquisition is also introducing a paradigm change in the way building blocks for vision algorithms are defined. One of the most fundamental and low-level building block is the feature definition and its extraction. Features are the visual inputs to many higher level algorithms such as tracking and other data association algorithms that themselves are building blocks to even more complex and larger algorithms.This work is carried out using a $640 \times 480$ pixels Asynchronous time-based image sensor but any event-based camera producing events can make use of this methodology that is independent from the sensor.  

\section{Related works}
Previous works that tackled the features detection problem account several corners detectors that are inferred from the optical flow computation~\cite{Clady2015}, or obtained by locally integrating events so the Harris operator can be computed~\cite{Vasco2016}. The time-surface as introduced in~\cite{Benosman_2014} is another way to achieve corner detection as it allows to define a compact spatio-temporal descriptor around an incoming event.  In~\cite{Mueggler2017}, the time-surface allows to build a support on top of which is applied the FAST~\cite{Rosten2006} detector. A variant to the FAST has also been introduced in~\cite{Alzugaray2018} to provide both detection and tracking. Another way to provide a spatial support for the corner detection is to make use of the DAVIS, which provides frame-on-demand. Corners and other usual features are detected on those frames by applying standard computer vision detectors, then tracking algorithms are operated on the events captured between the frames as it is proposed in~\cite{Tedaldi2016}.\\
An expectation-maximization approach, coupled with an iterative closest point algorithm is used in~\cite{Zhu2017} to keep track of corners that are detected at the beginning of the sequences from an edge map built by accumulating events over a manually selected integration time. Events are integrated over durations deduced from the optical flow and the edge map to generate unitary contours to then apply an ICP algorithm.

\section{Method}

\ssect{Structure as a motion invariant}
\label{movingobject}

\begin{figure}[h]
    \centering
    \includegraphics[width=\columnwidth]{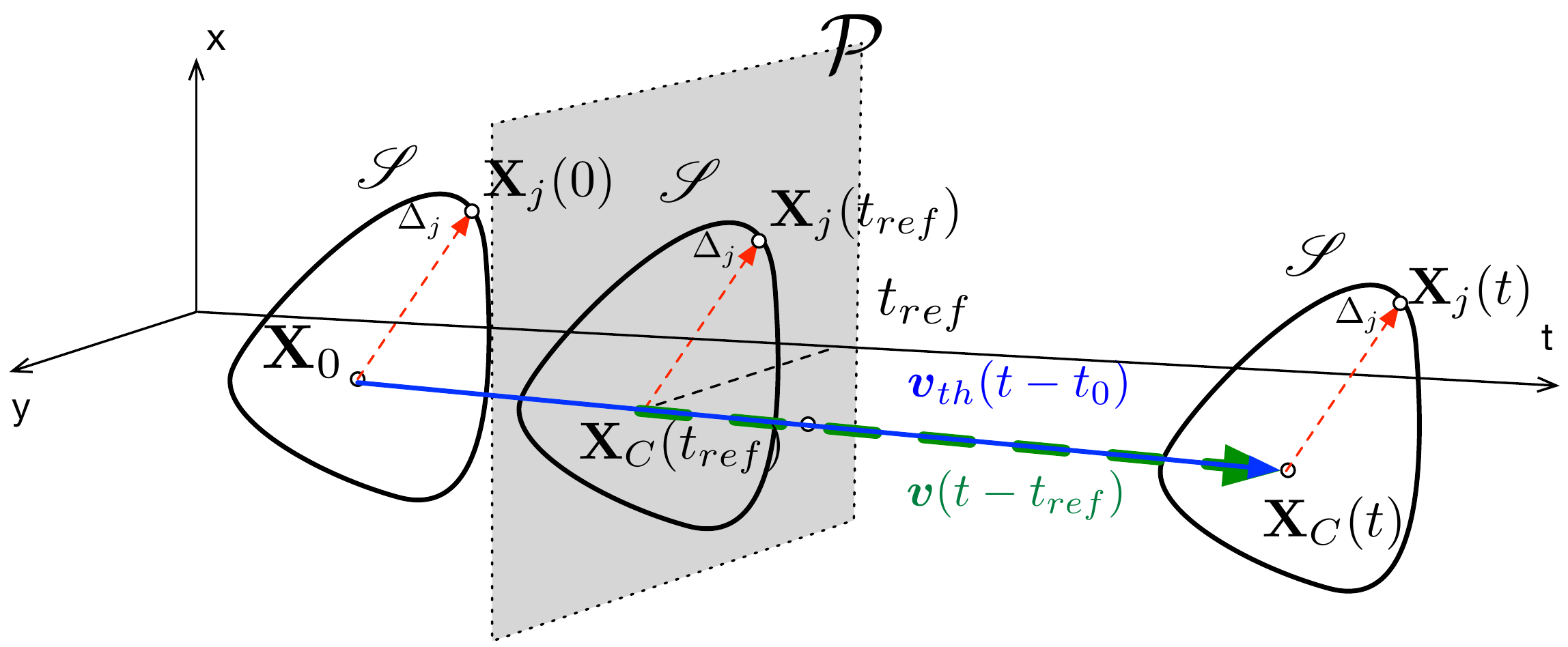}
    \caption{Object moving and generating the set $\mathscr{S}$ at constant velocity $\bm{v_{th}}$, between $t_0$ to $t$. $\bm{X_c}(t)$ is an arbitrary center of $\mathscr{S}$, at time $t$ and $\bm{\Delta_j}$ is a point of $\mathscr{S}$, expressed w.r.t. the center. The correct spatial distribution of $\mathscr{S}$ is recovered when $||\bm{v}-\bm{v_{th}}|| \rightarrow 0$. \label{3dproj}}
\end{figure}

Let $\mathscr{S}$ be the set of events generated by a moving object in the scene observed by the event-based camera. $\mathscr{S}$ is evolving in the space-time domain as shown in Figure~\ref{3dproj} and defined by:
\begin{equation}
\mathscr{S} = \left\{ \bm{X_j}(t) = \bm{X_{\mathrm{C}}}(t) + \bm{\Delta}_j \right\},
\end{equation}
with $\bm{X_{\mathrm{C}}}(t)$ an arbitrary center of reference of the observed object and $\bm{\Delta_j}$ the position of the $j^{th}$ point on the object w.r.t the center on a discrete grid, assuming the object is not deformable, at time $t$. When in motion, $\bm{X_j}(t)$ depends only on the new position $\bm{X}_C(t)$. If $\bm{X}_0 = \bm{X}_\mathrm{C}(t=t_0)$, and with the assumption that the motion of $\mathscr{S}$ has negligible rotational component, the point position at time $t$ can be written:

\[ \bm{X_j}(t) = \bm{X_0} + \bm{\Delta_j} + \bm{v_{th}} (t - t_0), \]
where $\bm{v_{th}}$ is velocity of the structure from $t_0$ to $t$, assumed to be constant within that time interval. To recover this generating contour of the observed object from the set of recorded events one needs to estimate accurately $\bm{v_{th}}$ such that we can achieve a description of $\mathscr{S}$ that is independent of the velocity. Namely once the correct direction and amplitude of motion inside the spatio-temporal space of event is determined, it becomes possible to estimate the generative contour of the oberved shape. 
Estimating $\bm{v_{th}}$ allows us to build a correct representation of $\mathscr{S}$ independent from the velocity vector.

\begin{figure}[h]
    \centering
    \includegraphics[width=0.49\textwidth]{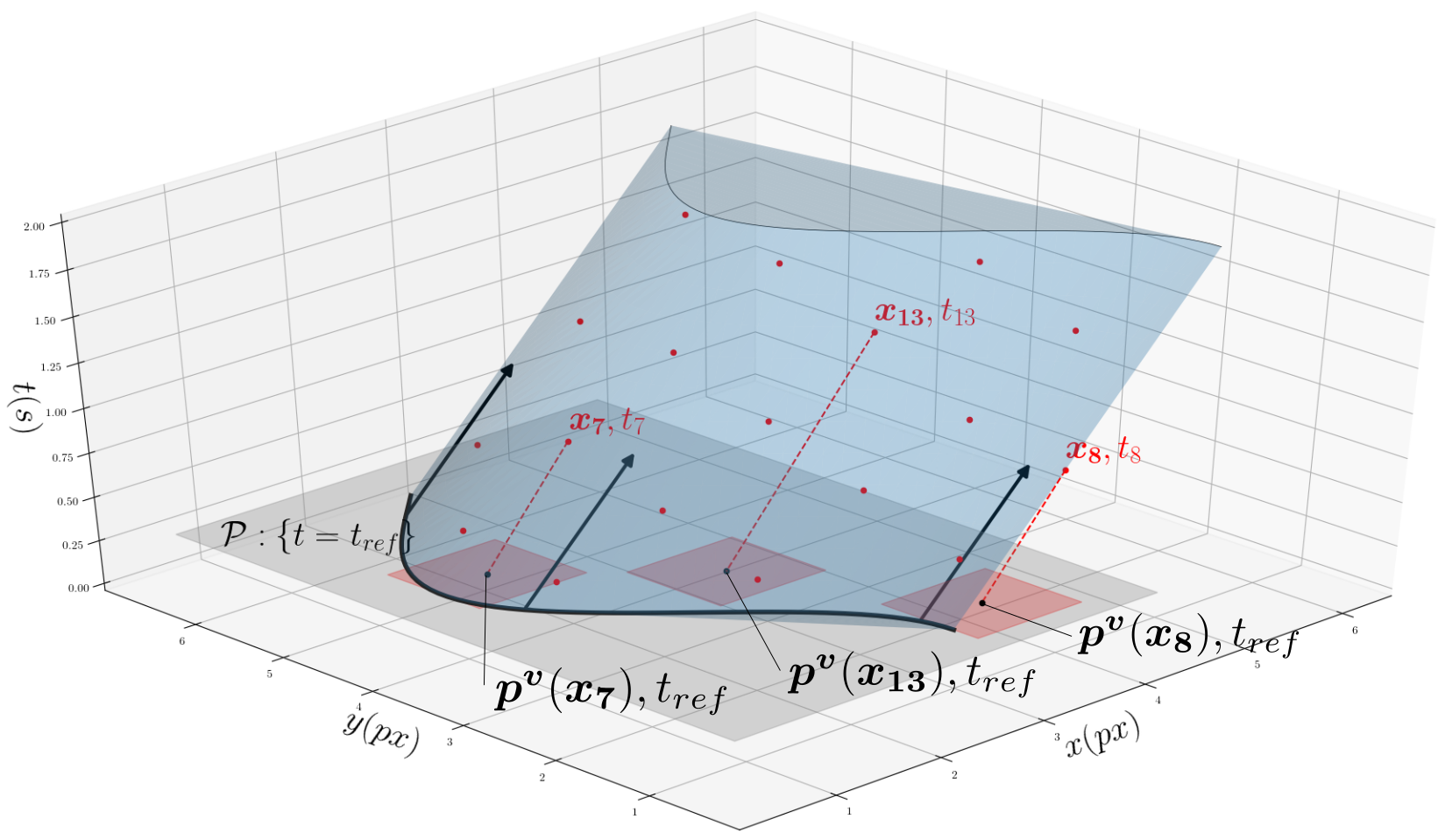}
    \caption{3D view of a spatio-temporal context, with events projected as described with $\bm{v_{th}} \neq \bm{v}$. \textit{Black thick line}: Local shape of the object. \textit{Black arrows}: Movement of the object. \textit{Red dots}: Events. \textit{Red dashes}: Projection of each event onto the $t_{ref} = 0.5s$ plan. \textit{Red squares}: Area affected by each projected event.}
    \label{3dprojseveral}
\end{figure}

Each event in $\mathscr{S}$ is generated at discrete time $t_i$ when the object is moving. For readability reason, we are dropping the index $j$ and the event $e_i = (\bm{X_j},t_i)$ is summarized into
\begin{equation}
\bm{x_i} = \bm{X_j}(t_i).
\label{eq:shorten}
\end{equation}
Considering the 3D spatio-temporal space, we denote by $\bm{v}$ the estimate of $\bm{v}_{th}$. We project each event at position $\bm{x}_i$ onto the plane $\mathcal{P}$ defined by $t=t_{ref}$ subject to $t_0 \le t_{ref} \le t$ (in practice, $t_{ref}$ is the time
at which we start to observe the structure and that does not necessarily needs to be equal to $t_0$).
The projection $\bm{p^v}(\bm{x_i})$ for an event on $\mathcal{P}$ is given by:
\begin{equation}
    \bm{p^v}(\bm{x_i}) = \bm{x_i} - \bm{v} (t_i - t_{ref}).
    \label{projectioneq}
\end{equation}
and represented in Figure~\ref{3dprojseveral} by the red dashed. Hence, according to equation~(\ref{eq:shorten}):
\begin{equation}
\begin{split}
        \bm{p^v}(\bm{x_i}) & = \bm{X_0} + \bm{\Delta_j} + \bm{v_{th}} (t_i - t_0) - \bm{v} (t_i - t_{ref}) \\
                  & = \bm{\Delta_j} + (\bm{v_{th}} - \bm{v}) t_i + \underbrace{(\bm{X_0} - \bm{v_{th}} t_0 + \bm{v} t_{ref})}_{\mu_0}
\end{split},
\label{eq:moving_set}
\end{equation}
$\mu_0$ is a time-independent quantity for $\mathscr{S}$ and the set $\{\bm{\Delta_j}\}$ defines the spatial distribution of $\mathscr{S}$. If $\bm{v}$ is correctly estimated, $\{\bm{\Delta_j}\}$ can be recovered by projecting the events generated by the moving object into $\mathcal{P}$,in the direction of $\bm{v}$, this is equivalent to have: 
\begin{equation}
    ||\bm{v_{th}} - \bm{v}|| \le \epsilon, 
    \label{speedcond}
\end{equation}
with $\epsilon$ arbitrarily small. This also means that the shape of the object will be reconstructed when the estimated speed is converging close enough to the theoretical one. With such consideration, we defined in this work an event-based feature as any stable, time-independent structure that can be recovered through observation. \\


\ssect{Spatial probability density function}
To establish the spatial distribution of $\mathscr{S}$, we introduce the quantity:

\begin{equation}
    \mathcal{U}_{\bm{v}}(x, y, t) = \sum \limits_{i, t_{ref} \leq t_i \leq t} \delta_{x, p^{\bm{v}}_x(\bm{x_i})} \delta_{y, p^{\bm{v}}_y(\bm{x_i})},
    \label{eq:histo_project}
\end{equation}
such that $p^{\bm{v}}_{x}(\bm{x_i})$ and $p^{\bm{v}}_{y}(\bm{x_i})$ are respectively the $x$ and $y$-components of $\bm{p^v}(\bm{x_i})$. With $\delta$ as the Kronecker function, the quantity $\mathcal{U}_{\bm{v}}$ is incremented by 1 at location $(x,y)$, each time an event occurs within the time interval $[t_{ref},t]$. (\ref{eq:histo_project}) defines an histogram along the direction of $\bm{v}$ as shown by Figure~\ref{3dprojseveral}: all events within the time interval $[t_{ref},t]$ and some spatial neighborhood, are contributing to the histogram.   
From this, after sufficient motion has been recorded, i.e. for $t$ such that $t-t_{ref}>>\frac{1}{||\bm{v}||}$, we can define the probability density function ($pdf$) as:  
\begin{equation}
    pdf_{\bm{v}}(x, y, t) = \frac{\mathcal{U}_{\bm{v}}(x, y, t)}{\max \limits_{x, y} \mathcal{U}_{\bm{v}}(x, y, t)}
    \label{eq:spdf}
\end{equation}

By thresholding this probability map with $\Pi \in ]0,1]$, one can recover the initial structure, defined here by
\begin{equation}
    \Gamma_{\bm{v}} = \{(x,y) | \lim \limits_{t \rightarrow \infty} pdf_{\bm{v}}(x, y, t) \geq \Pi\}
    \label{eq:threshold}
\end{equation}

The methodology is based on the high precision of the sensor and is making full use of its $\mu$s accuracy. Events can be projected in an almost continuous manner while estimating continuously the motion parameters. Typically, we can trade the high temporal precision for a much higher spatial accuracy. Thus, one can increase the spatial definition of $\Gamma_{\bm{v}}$ - initially given by the sensor's definition - by subpixeling $\mathcal{U}$ and $pdf_{\bm{v}}(x, y, t)$. The longer the structure with (\ref{speedcond}) satisfied is observed, the more accurate is the spatial description.\\
As $\Gamma_{\bm{v}}$ is defined through estimation of $\bm{v}$ and because we are using the currently available event-based vision sensors, we are facing one of the properties of signal acquisition: the sensor is blind to structures that are colinear to $\bm{v}$ since, along the edges, the luminance changes are negligible. For this reason, $\Gamma_{\bm{v}}$ still depends on the speed direction.

\ssect{Feature detection and tracking: spatiotemporal descriptor}
\label{featuretracking}

In this section, we focus on the tracking of local spatial structures that are, through the tracking mechanism, converging to stable projections. We rely on the previous general method of speed projection on a local spatial neighborhood $\mathcal{W}$ of size $R$, centered at some arbitrary position $\bm{X} $, for time $t=t_{ref}$.\\

A set of quantized velocities $\bm{v_i}$, in amplitude and in direction, are to be tested to estimate $\bm{v_{th}}$. Each projection along velocity $\bm{v_i}$ allows us to build the decaying map :
\begin{equation}
    \mathcal{D}_{\bm{v_i}}(x, y, t) = \sum \limits_{j, t_{ref} \leq t_j \leq t} \delta_{x, p^{\bm{v_i}}_x(\bm{x_j})} \delta_{y, p^{\bm{v_i}}_y(\bm{x_j})} e^{-(t-t_j)/\tau_i},
    \label{dmdef}
\end{equation}
with $\tau_i$ being a time constant specific to $\bm{v_i}$. 


From these maps, we can compute the mean spatial position in $\mathcal{W}$ of the events triggered by the moving object:
\begin{equation}
    \mean{\bm{X}_{i}}(t) = \begin{pmatrix}
        \frac{\iint_{\mathcal{W}} x \mathcal{D}_{\bm{v_i}}(x, y, t) dx dy}{\iint_{\mathcal{W}} \mathcal{D}_{\bm{v_i}}(x, y, t) dx dy}\\
        \frac{\iint_{\mathcal{W}} y \mathcal{D}_{\bm{v_i}}(x, y, t) dx dy}{\iint_{\mathcal{W}} \mathcal{D}_{\bm{v_i}}(x, y, t) dx dy}
    \end{pmatrix}
\end{equation}
by averaging the position of the projected events, weighted by the decaying kernel. This mean positions stands at the core of the tracking algorithm, as long as what is considered inside the observation window $\mathcal{W}$ presents a curvature radius at most of the size of the window. Such feature, on a low scale - a few pixels wide - can thus be a corner, part of a circle, segment of curve, \dots.\\

As stated in section \ref{movingobject}, we want to get the projection velocity $\bm{v_{i}}$ closest to the actual object speed $\bm{v_{th}}$ to create a stable structure. Thus, we need to detect and correct any displacement of the mean position of the projected events, that should remain still as long as these two speeds match. For this structure to be considered in its latest position, each of these projected events must thus disappear once the object has travelled a $1px$ distance, that is done in a typical time $\tau_i = \frac{1}{||\bm{v_i}||}$. \\

We can then estimate and store a reference position:
\begin{equation}
    \mean{\bm{X}}_{i, ref} \equiv \mean{\bm{X}_{i}}(t_{ref} + \tau_i).
    \label{refpos}
\end{equation}

The time reference $t_{ref} + \tau_i$ used means that this decaying map has been built over a time $\tau_i$, assuming the speed $\bm{v_i}$ was initially close enough from $\bm{v_{th}}$, all pixels reporting the feature's shape must have generated an event at least once.\\
From this reference, we can get an error in projection speed, for each event $e_i$, given by
\begin{equation}
    \bm{\epsilon_{i}}(t) = \frac{\mean{\bm{X}_{i}}(t) - \mean{\bm{X}}_{i, ref}}{t - t_{ref}}.
\end{equation}

Back propagating this speed error to the velocity $\bm{v_i}$, we can keep track of the feature. Using initial speeds distributed in the entire $(v_x, v_y)$ space allows for the feature to be tracked more easily, as the position reference $\mean{\bm{X}}_{i, ref}$ will be more accurate, and the error in speed smaller from the start. Adding a correction factor $\lambda_i(t) < 1$, at time $t_j$ of event $e_j$, we update $\bm{v_i}$ as:
\begin{equation}
    \bm{v_{i}}(t_{j+1}) = \bm{v_{i}}(t_j) + \lambda_i(t_j) \bm{\epsilon_{i}}(t_j),
\end{equation}
allows also to include inertia to the system, giving it a good resistance to noise, thus increasing the system stability. 


\ssect{Setting the correction factor $\lambda_i$}
\label{inertiasetting}
The value of $\lambda_i$ has to be set dynamically. Each event is an opportunity for the system to correct the projection speed $\bm{v_{i}}$. For highly dynamical features - complex shapes or high speed - a large number of events will be processed, and the system might become unstable and possibly oscillate around the theoretical speed value. To prevent that, one must set the $\lambda_i$ accordingly. To get an estimate of the number of events appearing for this feature, we can once again use the decaying map $\mathcal{D}_{\bm{v_i}}(x, y, t)$. 
Let us compute the sum over this decaying map: 
\begin{equation}
    \begin{split}
        S_i(t) &= \sum \limits_{(x, y) \in \mathcal{W}} \mathcal{D}_{\bm{v_i}}(x, y, t)\\
               &= \sum \limits_{(x, y) \in \mathcal{W}} \sum \limits_{j, t_j \leq t} \delta_{x, p^{\bm{v_i}}_x(\bm{x_j})} \delta_{y, p^{\bm{v_i}}_y(\bm{x_j})} e^{-(t-t_j)||\bm{v_i}||}\\
    \end{split}
    \label{sumdm}
\end{equation}
Summing over the patch means that we do not care about the actual position of the projected events, rather than just the length of the edges producing events. To simplify the computation, we change the notation to go from discrete to continuous time notation, with $\nu_{e, px}$ the event production rate of $\Gamma_{\bm{v}}$ for one pixel.
\begin{equation}
    \begin{split}
        S_i(t) &= \sum \limits_{(x, y) \in \mathcal{W}} \int_{t_{ref}}^t \nu_{e, px} e^{-(t-t')||\bm{v_i}||} dt'\\
               &= \sum \limits_{(x, y) \in \mathcal{W}} \frac{\nu_{e, px}}{ ||\bm{v_i}||} (1 - e^{-(t-t_{ref})||\bm{v_i}||})
    \end{split}
\end{equation}

Thus, for $(t - t_{ref}) \gg 1/||\bm{v_i}||$, that is true for a displacement of a few pixels, and defining 
\begin{equation}
    \nu_e \equiv \sum \limits_{(x, y) \in \Gamma_{\bm{v}}} \nu_{e, px} 
\end{equation}
the number of events produced per second by $\Gamma_{\bm{v}}$ at this particular velocity, we have
\begin{equation}
    S_i = \nu_e / ||\bm{v_i}|| 
\end{equation}
that is the number of events produced by $\Gamma_{\bm{v}}$ per second, divided by the projected speed. This $\nu_e$ event rate is the one that we will use to set the $\lambda_i$ value.
Now when we modify the velocity for each occurring event near our tracked feature, we have

\begin{equation}
    \label{speed_diff_eq0}
        \bm{v_i}(t_{j+1}) = \bm{v_i}(t_j) + \lambda_i(t_j) \bm{\epsilon_{i}}(t_j)
\end{equation}
and with a first order approximation, we get
\begin{equation}
    \label{speed_diff_eq}
        \frac{\partial \bm{v_i}(t_j)}{\partial t} (t_{j+1} - t_j) =  \lambda_i(t_j) \bm{\epsilon_{i}}(t_j)
    \end{equation}

On average, $(t_{j+1} - t_j)$ is equal to $1/\nu_e$, event rate of $\Gamma_{\bm{v}}$. Since $\bm{\epsilon_{i}}(t_j) = \bm{v_{th}} - \bm{v_i}(t_j)$, equation \ref{speed_diff_eq} finally gives us:  
\begin{equation}
        \bm{v_i}(t_j) = \bm{v_{th}} + (\bm{v_i}(t_{ref}) - \bm{v_{th}}) e^{(t_j-t_{ref}) \lambda_i \nu_e}
\end{equation}

We want the speed to converge after a displacement of only a few pixels $\Delta_x$, in a duration of $\Delta_x / ||\bm{v_i}(t_j)||$. Thus, assuming once more that $\bm{v_i}$ was initialized relatively close to $\bm{v_{th}}$:
\begin{equation}
    \begin{split}
        \Delta_x / ||\bm{v_i}(t_j)|| &= \frac{1}{\lambda_i(t_j) \nu_e} \\
                        \lambda_i(t_j) &= \frac{||\bm{v_i}(t_j)||}{\nu_e \Delta_x}\\
                                   &= \frac{1}{S_i \Delta_x}
    \end{split}
\end{equation}

This transformation allows to set the order of magnitude of the allowed displacement for the velocity to converge. We typically set $\Delta_x \sim R$, the size of the observation window $\mathcal{W}$.

\ssect{Features detection}

An easy reach would be to use the 2D content of the reference plane to apply a conventional frame-based feature descriptor. However, the use of a classical methodology from frame-based computer vision is not adequate with the pure event-based approach developed here and would imply operating on a 2D feature frame for each incoming event and unnecessary heavy computations. This could also restrict the tracking to self-defined features with known shapes, potentially reducing the tracking efficiency and possibilities. Also, we know that changing the observation direction of an object can change its shape, sometimes completely. Thus, what characterizes a feature is more a local information that is evolving over time, rather than a static description. The choice is then to let the algorithm ''decide'' which features to track, by monitoring its own variables to infer if the tracking is performing correctly. In this perspective, two variables are monitored. The first one is the ratio:
\begin{equation}
    A = \frac{S_i}{R}    
\end{equation}
with $S_i$ defined in equation \ref{sumdm} as the sum of the decaying map of velocity $\bm{v_i}$ within the observation window $\mathcal{W}$. The second one is the ratio between the norm of the error in speed and the norm of the velocity itself:
\begin{equation}
    B = \frac{||\bm{\epsilon_{i}}(t)||}{||\bm{v_i}(t)||}
\end{equation}
As each of these variables are computed to correct $\bm{v_i}$ for each event, it allows for a simple embedded features quality detection.
The second term provides a rough estimate of the correct tracking of a feature. The results given in the following section reports a value of $B < 1\%$ on average when a good tracking is performed. Still, another case can appear with $B = 0$ and yet no tracking, that is when the observation window has been initialized on a event-free part of the scene. We can also monitor variable $A$, assessing for the actual presence of dynamics in this part of the scene. We also can assume that:
\begin{equation}
        S_i \propto R \rho
\end{equation}
$\rho$ being the linear fill factor of $\Gamma_{\bm{v}}$ in the window of observation. For instance, $\rho \sim 1$ for regular shapes such as lines, corners, segments, \dots On the other hand, $\rho = 0$ for empty space, and $\rho \rightarrow R$ for moving textured objects.\\

\ssect{Descriptor}
The tracking uses the decaying maps presented in equation \ref{dmdef} as a descriptor. Using this descriptor, instead of binary frames or Time-Surfaces for example, allows to lower drastically the speed dependency. Indeed, each point in this descriptor can be seen - once normalized - as a probability density of the tracked feature to have this specific shape during its movement. Some of these projected pixels can remain hidden due to the aperture issue, but we fully use the potential of artificial retinas, as most other descriptors present either blur or inconsistent data due to the necessity for precisely fitted time constants, or positions history trails. We have seen in the previous section that the detection and tracking algorithms can hardly be separated : a detected feature is by essence a feature that has stabilized its projection speed, thus that can be tracked. This means that the descriptor used can report any sort of shape, not restricting itself to corners or specific features. This allowed to detect a wide variety of different shapes tracked during the different trials, some of them are shown in Figure~\ref{reconstruction}.\\
Finally, a probability density already suggests distances, such as the Bhattacharyya distance, to compare two of these descriptors. Such a comparison is important for future development, as comparing two descriptors is a necessary step for higher level tasks.

\begin{figure*}[ht]
    \centering
    \includegraphics[width=0.75\textwidth]{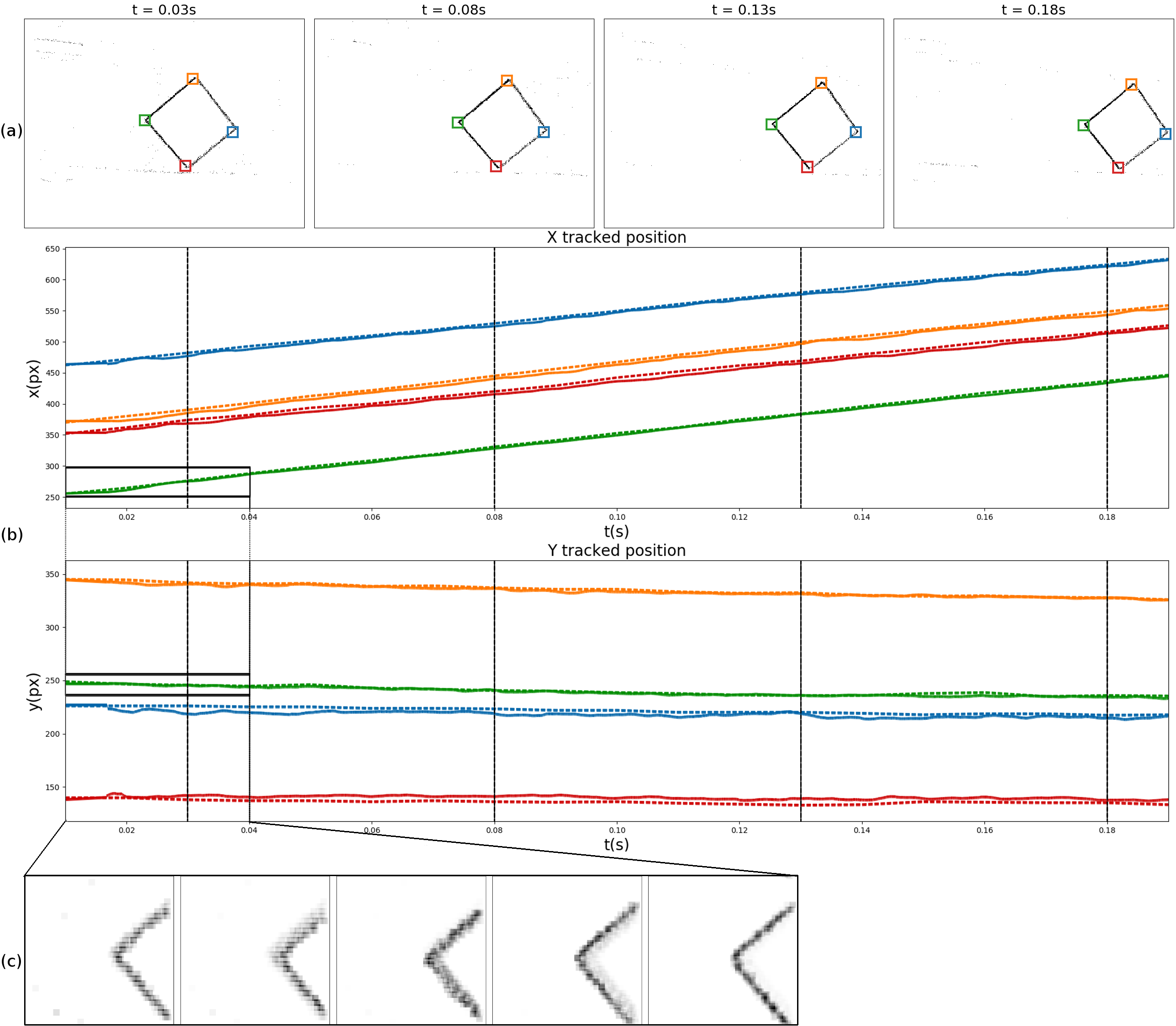}
    \caption{Tracking results of the four presented features. \textit{(a)} Frames created from the spatio-temporal context, with $5ms$ windows, reporting the visual correct tracking of each corner. \textit{(b)} $X$ and $Y$ positions of the tracked features, compared to the measured ground truth. A constant error can appear due to the time needed to compute the reference position $\bm{X_{i, t_0}}$ given in equation \ref{refpos}. \textit{(c)} Evolution of the projected feature $3$ at the early stages of the algorithm. The initial blurry shape disappears after a few milliseconds, showing the actual feature with enhanced precision as shown in Figure~\ref{reconstruction}.}
    \label{squaretracking}
\end{figure*}

\section{Implementations and performances}

\ssect{Tracking algorithm}

The estimated speed correction method follows the equations of sections \ref{featuretracking} and \ref{inertiasetting} summarized in Algorithm \ref{alg:speed-update-alg}.

\begin{algorithm}
    \caption{Speed Update}
    \label{alg:speed-update-alg}
    \begin{algorithmic}
        \While{event $e = \{\bm{x_e}, t_e\}$}
            \For{$\bm{v_{i}} \in \mathcal{V}_{est}$}
                \State $\bm{p_e^{v_i}} \gets \bm{x_e} - \bm{v_{i}} * (t_e - t_{0,i})$
                \If{$\bm{p_e^{v_i}}$ in observation box of speed $\bm{v_{i}}$}
                    \State Update $\mathcal{D}_i$
                    \State $S_i \gets \sum \limits_{x,y} \mathcal{D}_i(x,y)$
                    \State Compute $\mean{\bm{X}}_i$
                    \State $\bm{\epsilon_i} \gets (\mean{\bm{X}}_i - \mean{\bm{X}}_{i,ref}) / (t_e - t_{0,i})$
                    \State $\bm{v_i} \gets \bm{v_i} + \bm{\epsilon_i}/(S_i \Delta_x)$
                \EndIf
            \EndFor
        \EndWhile
    \end{algorithmic}
\end{algorithm}

Yet more elements have to be taken into account. The whole method is based on constant linear movements. If most trajectories can be approximated by several segments verifying this assumption, one must deal with the change between two of these segments. The method used in this work is described in Algorithm \ref{alg:plan-update-alg}.

\begin{algorithm}
    \caption{Projection Plan Update}
    \label{alg:plan-update-alg}
    \begin{algorithmic}
        \While{event $e = \{\bm{x_e}, t_e\}$}
            \State Update $\bm{v_i}$
            \If{$|\bm{\epsilon_i}| \leq k|\bm{v_i}|$ and $t_e - t_{0,i} > N/|\bm{v_i}|$ }
                \State $t_{0,i} \gets t_e - 1/|\bm{v_i}|$
                \State Update Observation box of $\bm{v_i}$
            \EndIf
        \EndWhile
    \end{algorithmic}
\end{algorithm}

We must verify two conditions to update the time of projection for each event. The first one is obviously that the norm of the error in speed computed $|\bm{\epsilon_i}|$ is small compared to the norm of the estimated speed itself $|\bm{v_i}|$, thus the feature is stable. We use in our case $k = 0.01$ that has been set experimentally.\\
The second condition is to ensure that a feature has moved sufficiently since the last projection on the reference plane. Although possible, it is better for computational efficiency to not systematically update the projection for each incoming event. Also, since the velocity is assumed to be correct with condition $1$ and the scene not changing at a MHz rate, we ask for a minimal displacement of $N$ pixels that has been set experimentally to $N=6$ as it provides the best match between stable tracking and real-time operation.

\ssect{Tracking corners at constant speed}


The first experiment will consider constant speed features tracking, travelling linearly. We assume, that a speed can be considered a continuous function of time with the time definition of the sensors used, this is a fair assumption as apart from specific phenomenon nothing in everyday scenes updates its dynamics at the MHz.\\
We recorded with the event-based camera \cite{ATIS} a fast moving $\ang{45}$-rotated square, whose ground-truth \textit{features} positions have been manually determined. We thus consider in this first example $4$ features, one for each corner, and apply the previously described tracking algorithm to the sequence. As we want here to check for the performances of the tracking part of the algorithm, we voluntarily disable the detection part of it. The initial speeds given to the algorithm are in a range of $-1000 px.s^{-1}$ to $1000 px.s^{-1}$ for both axis, for a ground truth of about $750 px.s^{-1}$. Results are shown in Figure \ref{squaretracking}.
As usual a good initialization improves the algorithm performances and allows it to converge faster. If some of the parameters necessary for its initialization have been found less important than expected, the most important one is the size of the observation window. This size must be greater than the typical size of an object we want to track, yet small enough to discriminate between different objects.

\begin{figure}[h!]
    \centering
    \includegraphics[width=0.47\textwidth]{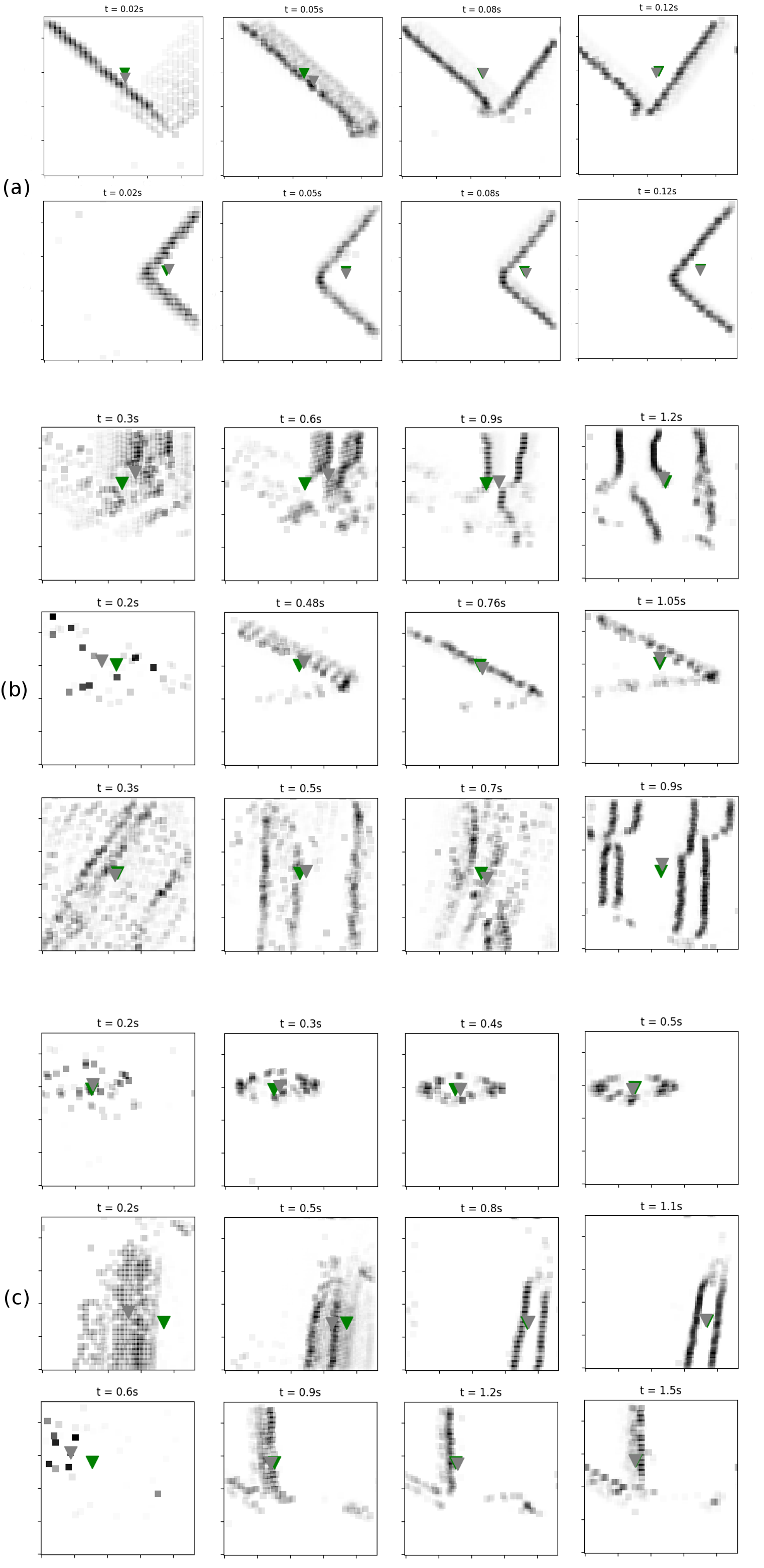}
    \caption{Non exhaustive list of features tracked by the algorithm. The green triangle gives the reference mean position stored for this feature, that the algorithm tries to stabilize, while the green triangle refers to the current mean position observed. (a) Two of the four features tracked the benchmark experiment presented in Figure \ref{squaretracking}. (b) Examples of features tracked in an outdoor environment. They can be, among others, gutter parts, manholes, or wall irregularities. (c) Examples of features tracked in an indoor environment. The observed size of the objects allows for an improved efficiency in detecting and tracking, and widens the variety of features, such as ceiling spotlights, window outline, or furniture edges.}
    \label{reconstruction}
\end{figure}

\ssect{Indoor and outdoor natural scenes}

\begin{figure*}[h!]
    \centering
    \includegraphics[width=0.80\textwidth]{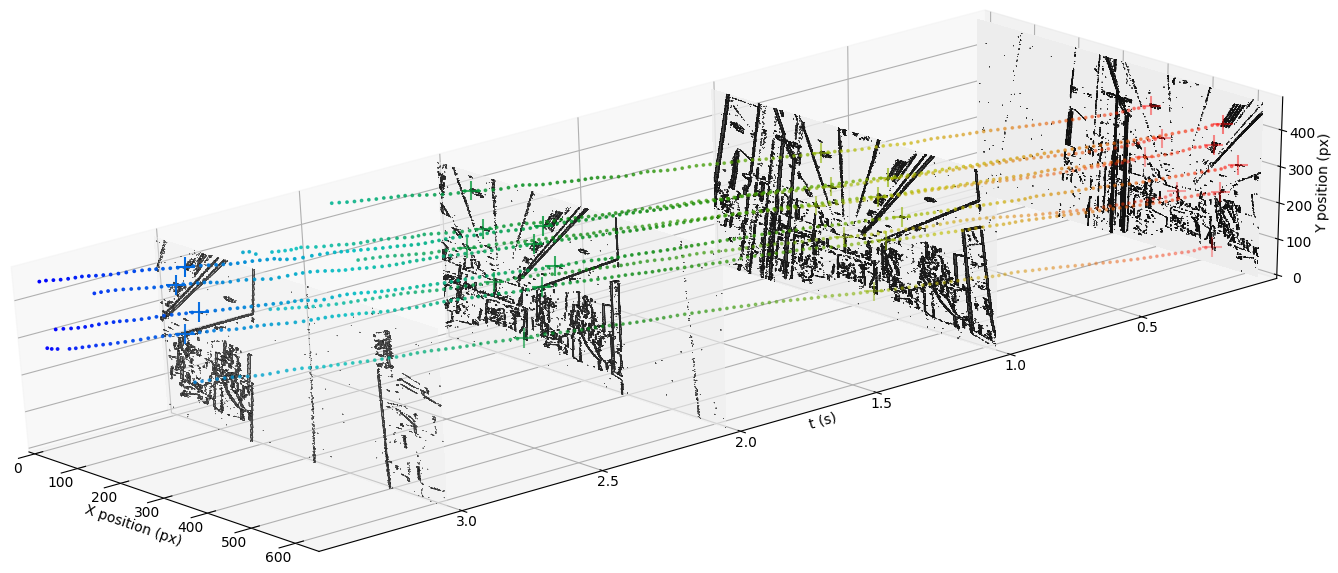}
    \caption{The indoor experiments have shown great stability in tracking, with a wide variety of features. Projecting this tracking in a 3D space allows to easily recover the movement from which originated the events.}
    \label{3D-office-figure}
\end{figure*}

 \begin{table}[h]
\resizebox{\columnwidth}{!}{
\begin{tabular}{|l|c|c|c|c|r|}
       \hline
       Scene & $N_{points}$ & $\mean{\epsilon}$ & $\mean{L}$ & $\mean{\epsilon / L}$ & $\sigma_{\epsilon / L}$\\
       \hline
       \hline
       Indoor & & & & & \\ 
       Rotation & $7$ & $2.4px$ & $382.4px$ & $1.4\%$ & $1.2\%$ \\
       \hline
       Outdoor & & & & & \\ 
       Rotation & $4$ & $4.1px$ & $417.9px$ & $1.0\%$ & $0.25\%$ \\
       \hline
       Street  & & & & & \\ 
       Fixed Cam. & $2$ & $8.6px$ & $470.2px$ & $1.8\%$ & $0.07\%$ \\
       \hline
\end{tabular}}
    \caption{Results of the tracking algorithm in different environment. We only consider the first set of points tracked for each scene for clarity purposes. We propose the absolute errors and tracking lengths $\epsilon$ and $L$ averaged over each set, and the average of a normalized error $\epsilon / L$. The comparison between the tracking results and the ground truth starts once the followed features enter the observation box - usually a $30px$-sided square.}
    \label{results-table}
\end{table}    
    
Results shown in Table \ref{results-table} report results for all trackers for the whole sequence. 
The number of points considered $N_{points}$ is given, with the average error $\mean{\epsilon}$ and the average length of tracking $\mean{L}$. Finally, we propose a normalized error $\mean{\epsilon / L}$, that can be related to a drift of the feature with respect to travel length. 
A good sample of the different shapes tracked during these experiments are reported in Figure \ref{reconstruction}. In this panel, we display the convergence over time of several features, stabilizing toward recognizable objects, such as spotlights, manholes, doors, \dots.
The indoor tracking is represented in a $3D$ space in Figure \ref{3D-office-figure}, and helps visualize the convergence of several independent tracked features towards coherent movement speeds.\\

These results emphasize the accuracy of the method. If the largest relative errors peak at around $3\%$, it is usually closer to $1\%$ all along the trajectory. The first (indoor) scene and the second (outdoor) scene show that the method is operating reliably despite scenes having been recorded with a hand-held camera, without any stabilization. The saccadic motion induced by the hand has little impact on the tracking performance. However, the third scene, captured by a static camera observing pedestrians is more challenging, because the local linear motion hypothesis is harder to fulfill: the pedestrians head motions are mainly made of large amplitude vertical oscillations. But the algorithm managed to track reliably such complex features. These performances are to be assessed w.r.t. to the task we want to achieve. Additional materials provide videos of each experiment.



\section{Conclusion}

    This work introduced an event-based method to track and estimate the velocity of features while enabling a native bayesian description of the feature. The method can detect and track non-specific and non specified features at different velocities and in different observation conditions. Tracking benchmarks have shown that the method is accurate while operating on each incoming event. This has been made possible by estimating reliable velocities for each tracked feature. Most importantly, using several projection velocities for initialization allowed to have very little dependency to the tuning parameters. The variety of features tracked show the potential of this type of algorithm, allowing for precise tracking of features with few constraints. The use of different feedback mechanisms should improve even more the stability of the method. Beyond the presented application of tracking and being able to describe features, this paper sets a general framework for event-based cameras by introducing a scheme of intertwined computation between space and time that makes use of the precise timing of each incoming event. The relation between space and time is made possible by the high temporal resolution of these sensors. This property allows for a natural writing of visual dynamics in the velocity domain. Velocity is surely the most interesting, reliable and straightforward elementary information one can reliably extract from these sensors.


\bibliographystyle{IEEEtran}


\end{document}